\documentclass[]{interact}
\usepackage{siunitx}
\usepackage{epstopdf}
\usepackage{subfigure}

\theoremstyle{plain}

\theoremstyle{definition}

\theoremstyle{remark}

\begin{document}


\title{Large-scale Weakly Supervised Learning for Road Extraction from Satellite Imagery}

\author{
\name{Shiqiao Meng\textsuperscript{a}, Zonglin Di\textsuperscript{b}, Siwei Yang\textsuperscript{c} and Yin Wang\textsuperscript{c}\thanks{Yin Wang. Email: yinw@tongji.edu.cn}}
\affil{\textsuperscript{a}The State Key Laboratory of Disaster Reduction in Civil Engineering, Tongji University, Shanghai, China; \textsuperscript{b}The College of Electrical and Computer Engineering, University of California, San Diego, United States; \textsuperscript{c}The College of Electronic and Information Engineering, Tongji University, Shanghai, China}
}
\maketitle

\begin{abstract}
Automatic road extraction from satellite imagery using deep learning is a viable alternative to traditional manual mapping.  
Therefore it has received considerable attention recently.
However, most of the existing methods are supervised and require pixel-level labeling, which is tedious and error-prone.
To make matters worse, the earth has a diverse range of terrain, vegetation, and man-made objects.
It is well known that models trained in one area generalize poorly to other areas.
Various shooting conditions such as light and angel, as well as different image processing techniques further complicate the issue.
It is impractical to develop training data to cover all image styles.
This paper proposes to leverage OpenStreetMap road data as weak labels and large scale satellite imagery to pre-train semantic segmentation models. 
Our extensive experimental results show that the prediction accuracy increases with the amount of the weakly labeled data, as well as the road density in the areas chosen for training.
Using as much as 100 times more data than the widely used DeepGlobe road dataset, our model with the D-LinkNet architecture and the ResNet-50 backbone exceeds the top performer of the current DeepGlobe leaderboard. 
Furthermore, due to large-scale pre-training, our model generalizes much better than those trained with only the curated datasets, implying great application potential.
\end{abstract}


\section{Introduction}
Accurate road maps play an essential role in the travel of our modern lives. As necessary support for autonomous driving, there is an increasing demand for higher-precision and higher-real-time mapping. The traditional method of making road maps requires land surveying, manual editing, and verification, which is costly, error-prone, and not real-time. 

With the development of deep learning technology in various computer vision tasks (such as image semantic segmentation), CNN-based methods can consistently achieve better results than traditional machine learning algorithms (Krizhevsky, Sutskever and Hinton 2012). This breakthrough has also inspired many researchers to use CNN-based models to extract roads from satellite images automatically (Zhou, Zhang and Wu 2018; Buslaev et al. 2018). Although this automated road extraction algorithm can achieve real-time extraction of roads in satellite images, a small part of its prediction errors will bring significant challenges to automatic mapping. A better way to improve model prediction accuracy is to train the model with a large amount of data in advance so that the pre-trained model can be trained on a small amount of data and perform higher accuracy. Such training methods are commonly used in various computer vision tasks (He, Girshick and Doll{\'a}r 2019; Zoph et al. 2020). 
In image classification tasks, previous researchers have proved through experiments that pre-training with multiple orders of magnitude larger than standard datasets containing incorrect labels can still bring a substantial increase in inaccuracy (Mahajan et al. 2018).
However, there is no such work for the semantic segmentation task because it is tough to get the label even the label is of low quality.

Unlike the typical semantic segmentation, the existing map information can provide the label for road extraction somehow, which provides the possibility of getting the label for the large-scale weakly supervised training.
OpenStreetMap (OSM) maps are drawn based on handheld GPS devices, aerial photographs, satellite images, and other free content, which Steve Kester created in July 2004 (Haklay and Weber 2018).
The OSM data contains the location information of roads in most parts of the world, so its data volume can be many orders of magnitude larger than that of other satellite road datasets. 
Since the storage method of roads in OSM data stores their rough positions rather than accurate road geometric edges, OSM data can be used as weakly supervised data to pre-train the model after image preprocessing.



This paper proposes a method to process OSM data, which converts road position information into masks that can be used for segmentation, thereby generating weakly supervised labels with different data volume scales for road extraction from satellite images. We proved that the versatility of these methods by using D-LinkNet(ResNet-18 backbone), D-LinkNet(ResNet-34 backbone), D-LinkNet(ResNet-50 backbone) (Zhou, Zhang and Wu 2018), and a deeper U-Net (Ronneberger, Fischer and Brox 2015) to pre-train on our large-scale dataset. And then, we use transfer learning to test on two open-source datasets (DeepGlobe road extraction dataset (Demir et al. 2018) and Mnih Massachusetts roads dataset (Mnih 2013)), which verified that pre-training with a large amount of OSM data could bring a substantial increase in accuracy of the prediction results. Moreover, we quantitatively verified that as the volume of data increases, the effect of the model would perform better.
We also studied the effect of fixing part of the model's parameters when performing transfer learning on the results through experiments.
In addition, we conducted an ablation study on the impact of road density in the pre-training dataset on transfer learning by selecting three densities of pre-training datasets that contains various scales.

To summarize, our contributions are:

$\bullet$ A method of processing OSM data is proposed, which can convert OSM data into labels that can be used for road recognition in satellite images, thereby making large-scale weakly supervised satellite image segmentation possible. 

$\bullet$ Using D-LinkNet with a ResNet-34 backbone, we conducted quantitative experiments on our self-made datasets, which are several orders of magnitude larger than the existing datasets, and verified the effectiveness of using OSM data to improve the mIoU of the predicted results. 
At the same time, a quantitative study was carried out on the relationship between the volume of the dataset and the improvement effect. 

$\bullet$ We test the impact of road density, fine-tuning methods, and the model capacity on the predicted results and provide novel empirical data by conducting ablation studies.

The remainder of this paper is as follows. 
Section \ref{sec:related} presents the related works. 
Section \ref{sec:prop}  gives a detailed description of the proposed methods. 
Section \ref{sec:exp} shows the datasets, experiments, implementation details, ablation studies and the results and Section \ref{sec:con} concludes the paper.


\section{Related Work}
\label{sec:related}
In order to solve the problem of road extraction from satellite images, researchers have proposed plenty of approaches. Mnih and Hinton (2012) are the first researchers to propose a CNN-based method for road extraction. 
Bastani et al. (2018) proposed RoadTracer, which uses an iterative search process guided by a CNN-based decision function to derive the road network graph directly from the output of the CNN.
Zhang et al. (2019) proposed an end-to-end method using a generative model to extract roads in images. Zhou, Zhang and Wu (2018) proposed a semantic segmentation model called D-LinkNet, which adopts encoder-decoder structure, dilated convolution, and pre-trained encoder for road extraction tasks and is efficient in computation and memory. M{\'a}ttyus, Luo and Urtasun (2017) used a semantic segmentation model based on deep learning to obtain the initial satellite road segmentation results and then proposed an algorithm that reasons about missing connections in the extracted road topology as the shortest path problem that can be solved efficiently. Zhang, Liu and Wang (2018) proposed a deep residual U-Net, which combines the strengths of residual learning and U-Net to extract roads in satellite images. Shan and Fang (2020) proposed a model that employs ResNet-18 and Atrous Spatial Pyramid Pooling technique to produce a model, and a point rend algorithm is used to recover a smooth and sharp road boundary. Besides, they proposed a modified cross-entropy loss function to train their model. 
Wang, Seo, and Jeon (2021) proposed an efficient nonlocal LinkNet with nonlocal blocks that can grasp relations between global features.
Wan et al. (2021) proposed a DA-RoadNet which can effectively solve discontinuous problems and preserve the integrity of the extracted road.

To improve the efficiency and accuracy of road extraction, some researchers have proposed methods that use other types of data to assist image data as input to improve the model's effectiveness. Sun, Di and Wang (2018) proposed a method to combine satellite imagery with GPS data to improve road extraction quality and outperform the RGB-only model by nearly 13$\%$ on mean IoU. Sun et al. (2019) proposed a method to leverage crowdsourced GPS data to improve and support road extraction from aerial imagery through novel data augmentation, GPS rendering, and 1D transpose convolution techniques, which show almost 5$\%$ improvements over previous competition winning models, and much better robustness when predicting new areas without any new training data or domain adaptation.

Some researchers proposed methods for weakly supervised learning using OSM data. 
Wu et al. (2019) proposed a novel weakly supervised method to extract road networks from very high resolution (VHR) images using only the OSM (OpenStreetMap) road centerline as training data. 
Pan, Zhang and Zhang (2021) proposed a generic and automatic approach for extracting road networks from VHR remote sensing images based on a fully convolutional neural network, in which the road centerlines from OSM have been employed to construct the labels for the model training and validation, and conducted experiments on various VHR image datasets with two different spatial resolutions of 0.3 and 1 m. Bonafilia et al. (2019) proposed using OSM data to produce weakly supervised data, then using the produced one million images for training on the D-LinkNet (ResNet-34 backbone) model, and compared their globally trained model with a model trained on the DeepGlobe regional dataset.

\section{Proposed Method}
\label{sec:prop}
\begin{figure}
    \centering
    \includegraphics[width=1.0\textwidth]{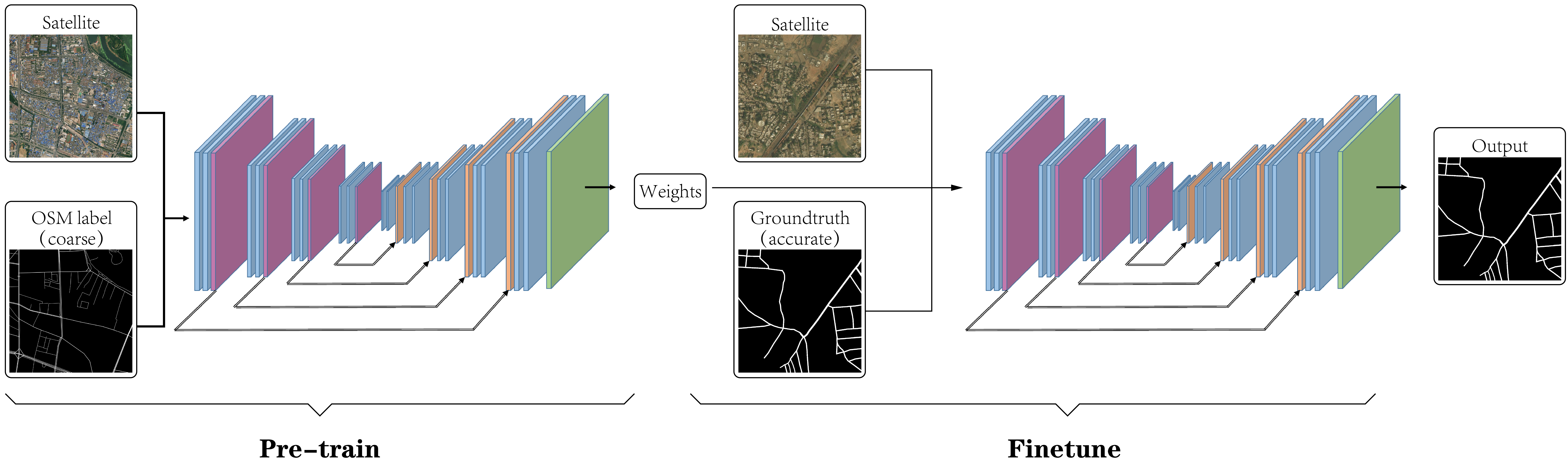}
    \caption{The framework of our methods. We pre-train the model using aerial images worldwide and the mask rendered by the OSM data. Then, we fine-tune the pre-trained model using the target training set with an accurate mask, leading to a better result.}
    \label{fig:framework}
\end{figure}

The framework of the proposed method is shown in Fig. \ref{fig:framework}. We first use a large number of aerial images in 18 zoom levels obtained from the Internet and the label data generated from OSM data to pre-train the model. These labels are relatively rough compared to the ground truth. 
Then, we use the weights obtained from the pre-trained model as the initial parameters and perform transfer learning by using a small dataset labeled manually to obtain the final model.



\subsection{Pre-training Dataset}

\subsubsection{Aerial Image Dataset}
\label{sec:aerial}
\begin{figure}
    \centering
    
    \includegraphics[width=\columnwidth]{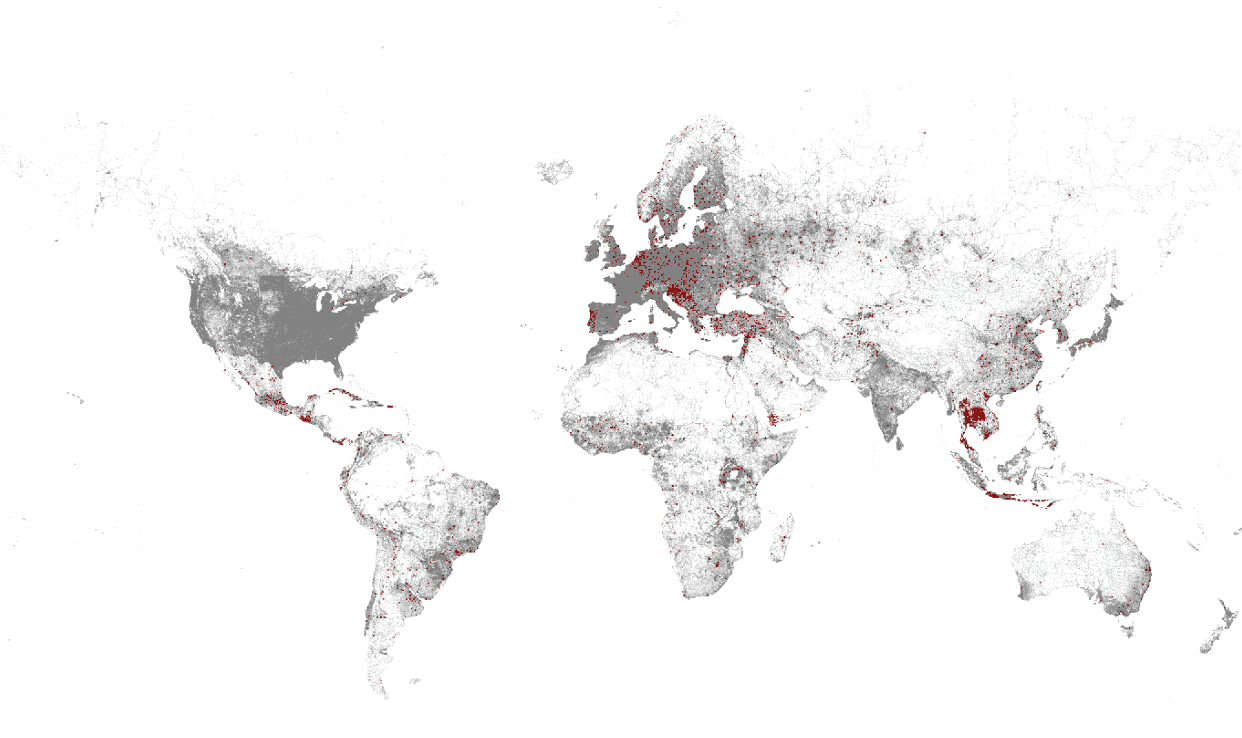}
    \caption{The distribution of the images we used around the world. The black point means at least one road in the corresponding tile in the OSM data, while the red point is the tile image we collected. Most images we have collected are from the area with the highest road density.}
    \label{fig:heatmap}
\end{figure}

To make the pre-train dataset as large as possible, we collect the data from over 200 countries. 
All the images are in 18 zoom levels, which means 1 pixel stands for 0.5 meters in the real world.
The distribution of the tiles we use around the world is shown in Fig. \ref{fig:heatmap}.
To improve efficiency, we stitch 256 original tiles, which are $256\times256$ pixels, into a $4096\times4096$ pixels image.
We also apply a random crop data augmentation to improve the generalization of the model using the much larger size of the stitched images than the size of each tile and the input of the network, which would be introduced in Section \ref{subsec:random_sample}.

\subsubsection{OSM Label Rendering}
\begin{figure}[htbp]
    \centering
    \subfigure[]{
        \includegraphics[width=0.21\columnwidth]{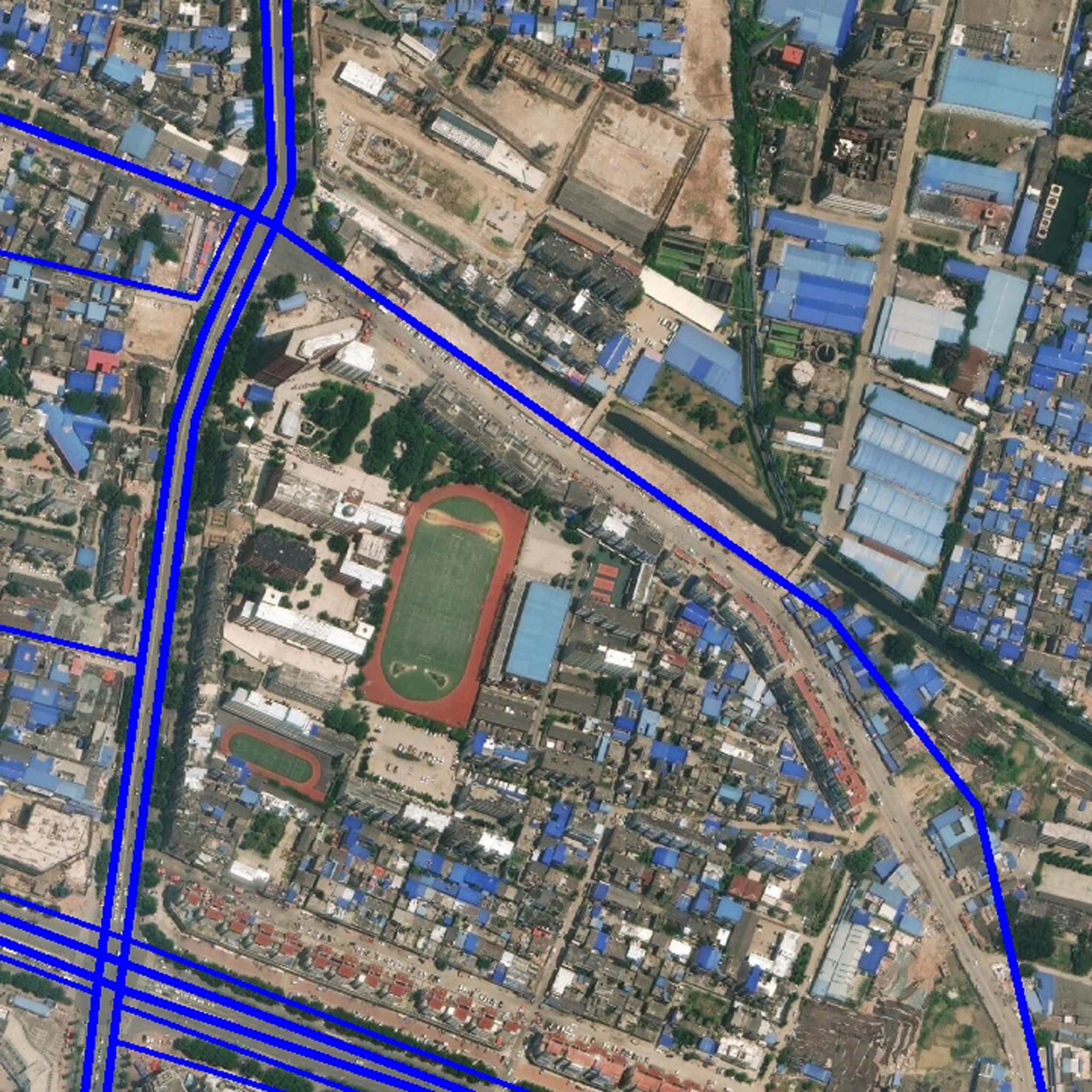}

    }
    \subfigure[]{
	\includegraphics[width=0.21\columnwidth]{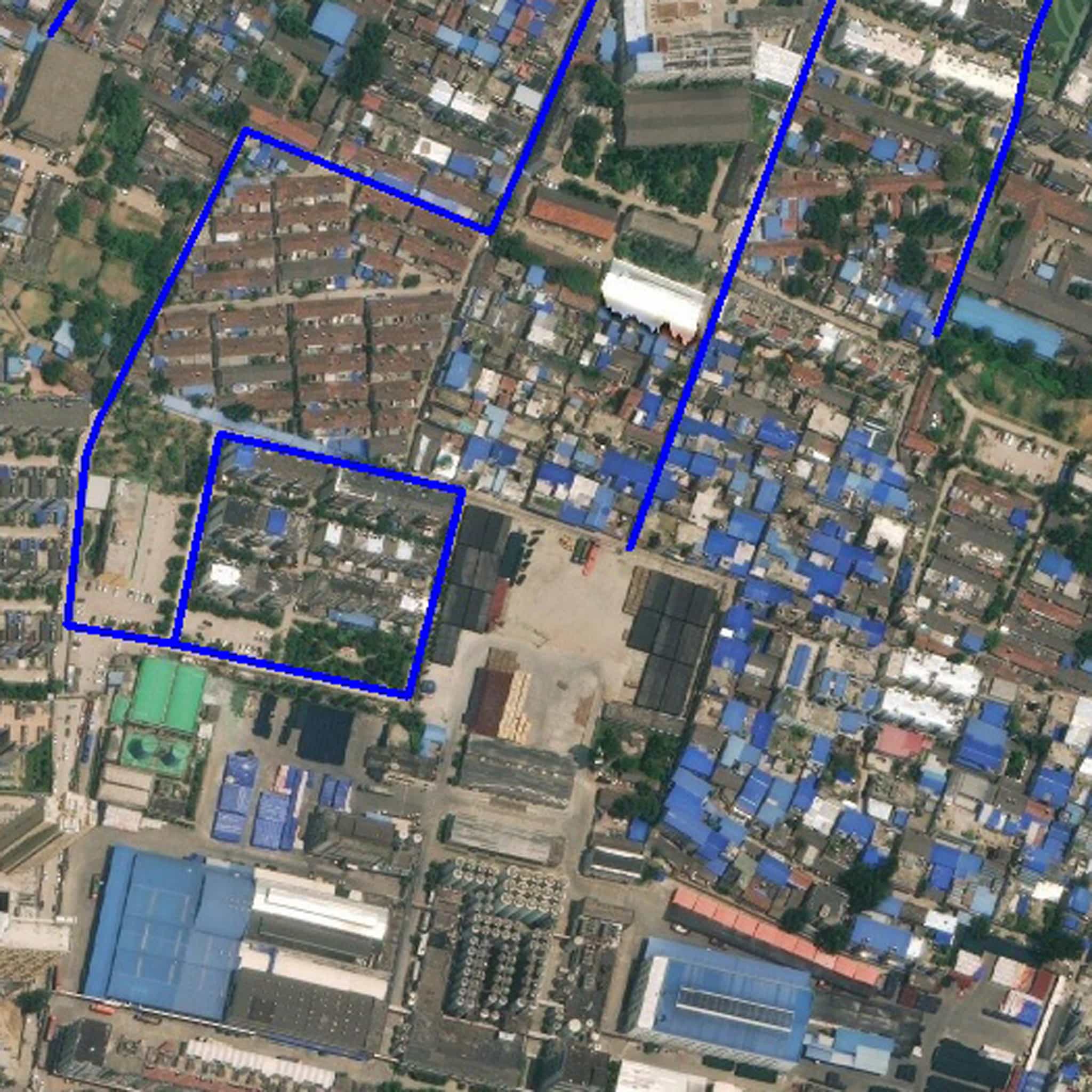}
    }
    \quad    
    \subfigure[]{
    	\includegraphics[width=0.21\columnwidth]{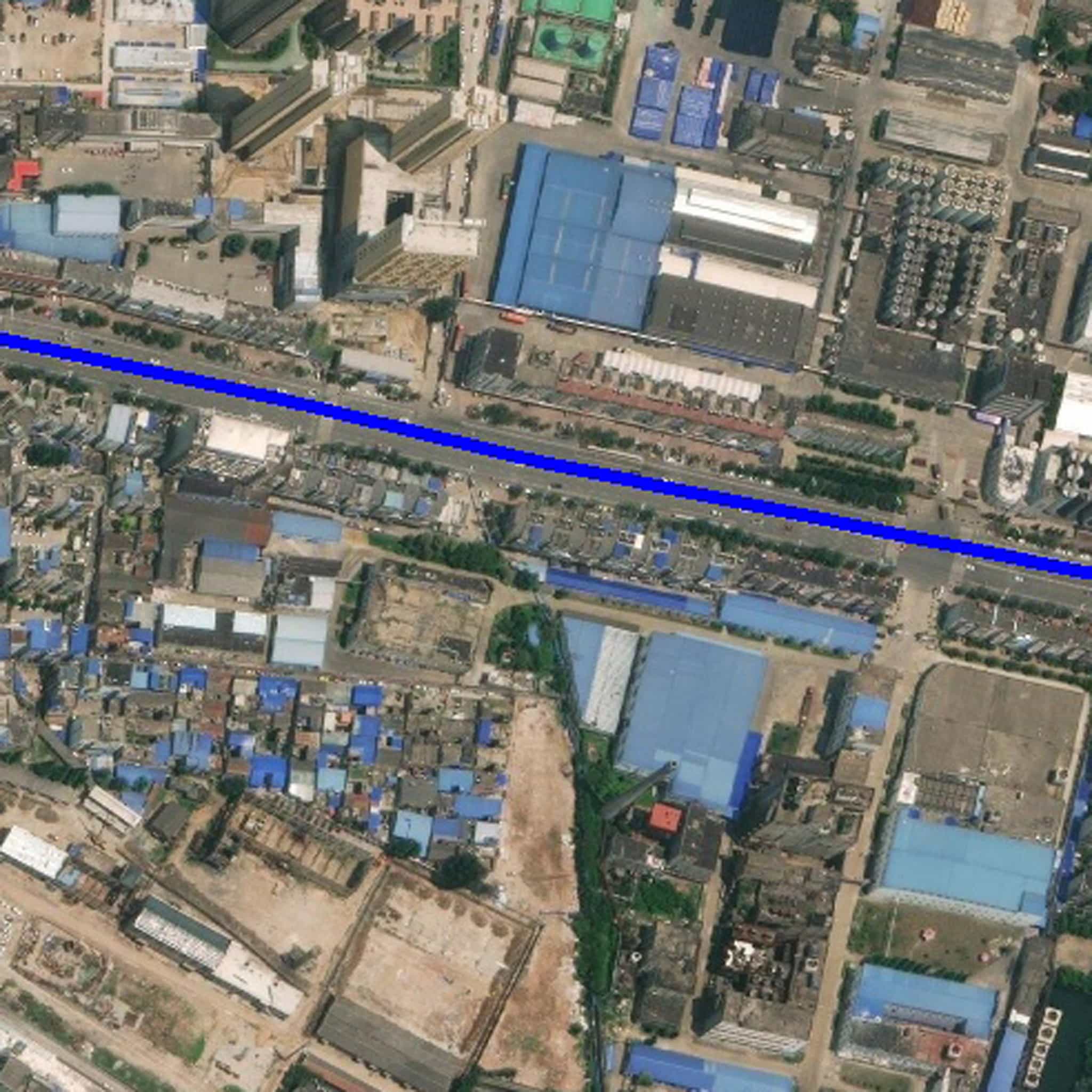}
    }
    \subfigure[]{
	\includegraphics[width=0.21\columnwidth]{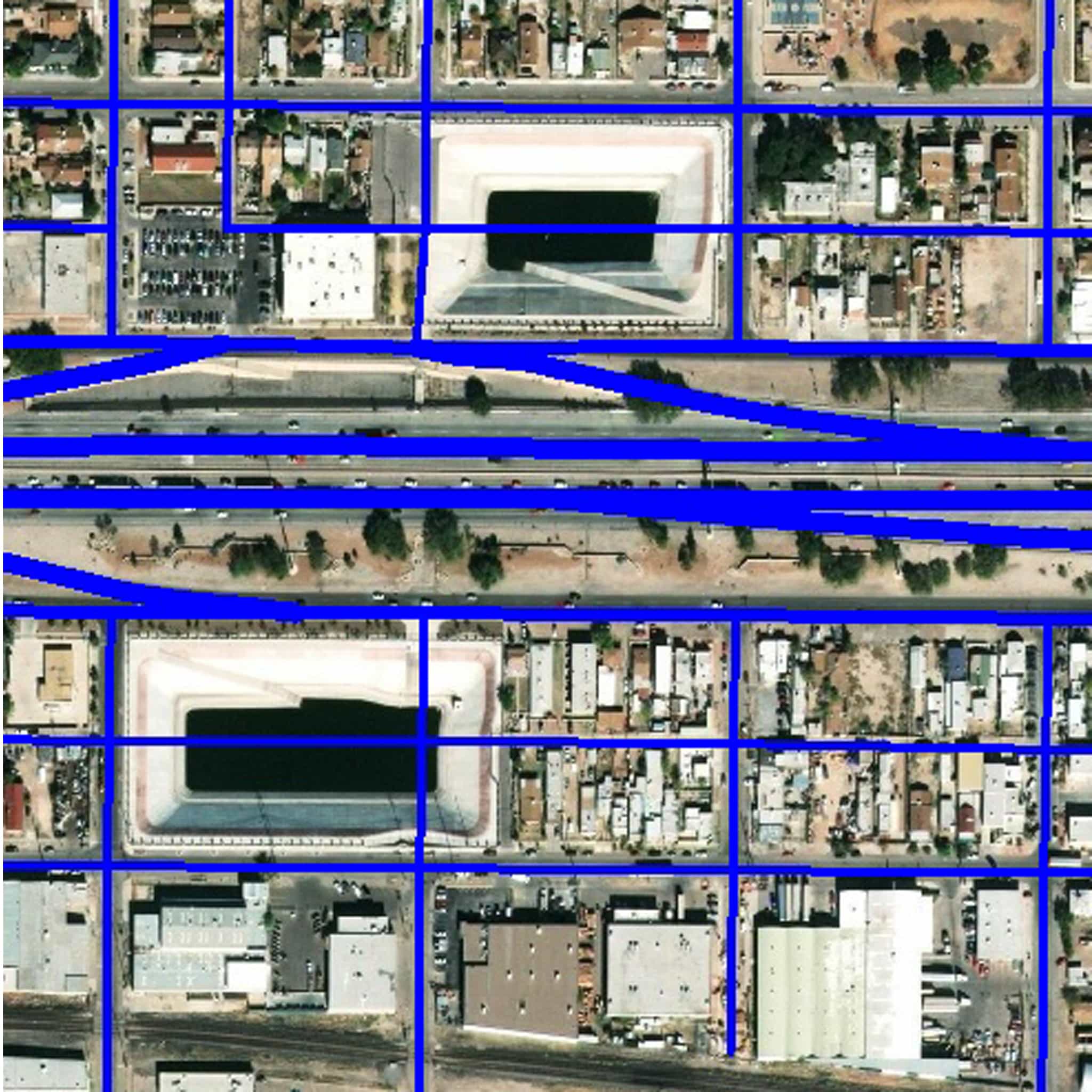}
    }
    \caption{Illustration of flaws in the masks rendered by OSM labels. In Fig. \ref{fig}(a), there is an offset between the rendered label and the actual road. In Fig. \ref{fig}(b), some of the labels are missing for the road between buildings. In Fig. \ref{fig}(c), the rendered label is not wide enough to cover the whole road. In \ref{fig}(d), the label is not accurate, so that the label crosses the building. All the examples show that the OSM dataset is far from an accurate dataset.}
\label{fig}
\end{figure}

We render the label from the global OSM data.
The original global OSM data is in XML format.
We extract the large, global OSM file into several small files according to the longitude and latitude boundary of the stitched satellite images mentioned above so that it would be faster to render the segmentation annotations.
Among the road label in the OSM label, we only focus on the paved roads. 
We render the roads according to their types with three different widths.
We render the main roads using 15-pixel lines, including motorway, motorway link, trunk and trunk link.
The middle roads are plotted in 10-pixel lines, including primary, primary link, secondary, secondary link, tertiary and tertiary link.
The rest small roads are labeled in 5-pixel lines.
Due to errors in OSM data and the shortcomings in the rendering process, there are many flaws in the generated labels, which are shown in Figure \ref{fig}.

\begin{figure}[htbp]
    \centering
    \includegraphics[width=10cm]{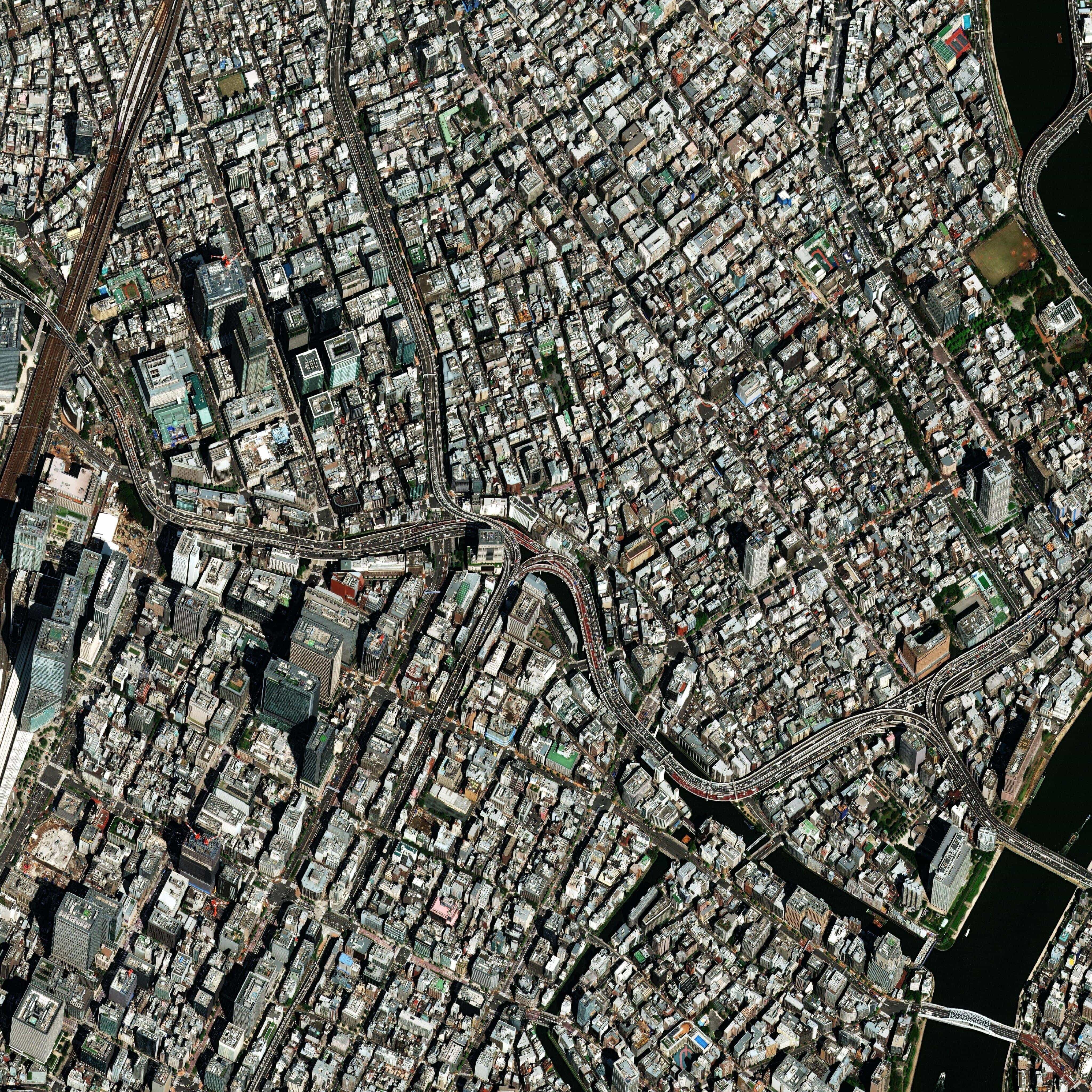}\hspace{5pt}

    \caption{The image and the rendered label have the highest density is from \ang{35;41;}N, \ang{139;46;}E (Tokyo, Japan) and its density is 24.107\%.}
    \label{fig:most}
\end{figure}

\subsection{Road Density and Pre-training Dataset Size}
\begin{figure}[!htbp]
    \centering
	\includegraphics[width=0.5\columnwidth]{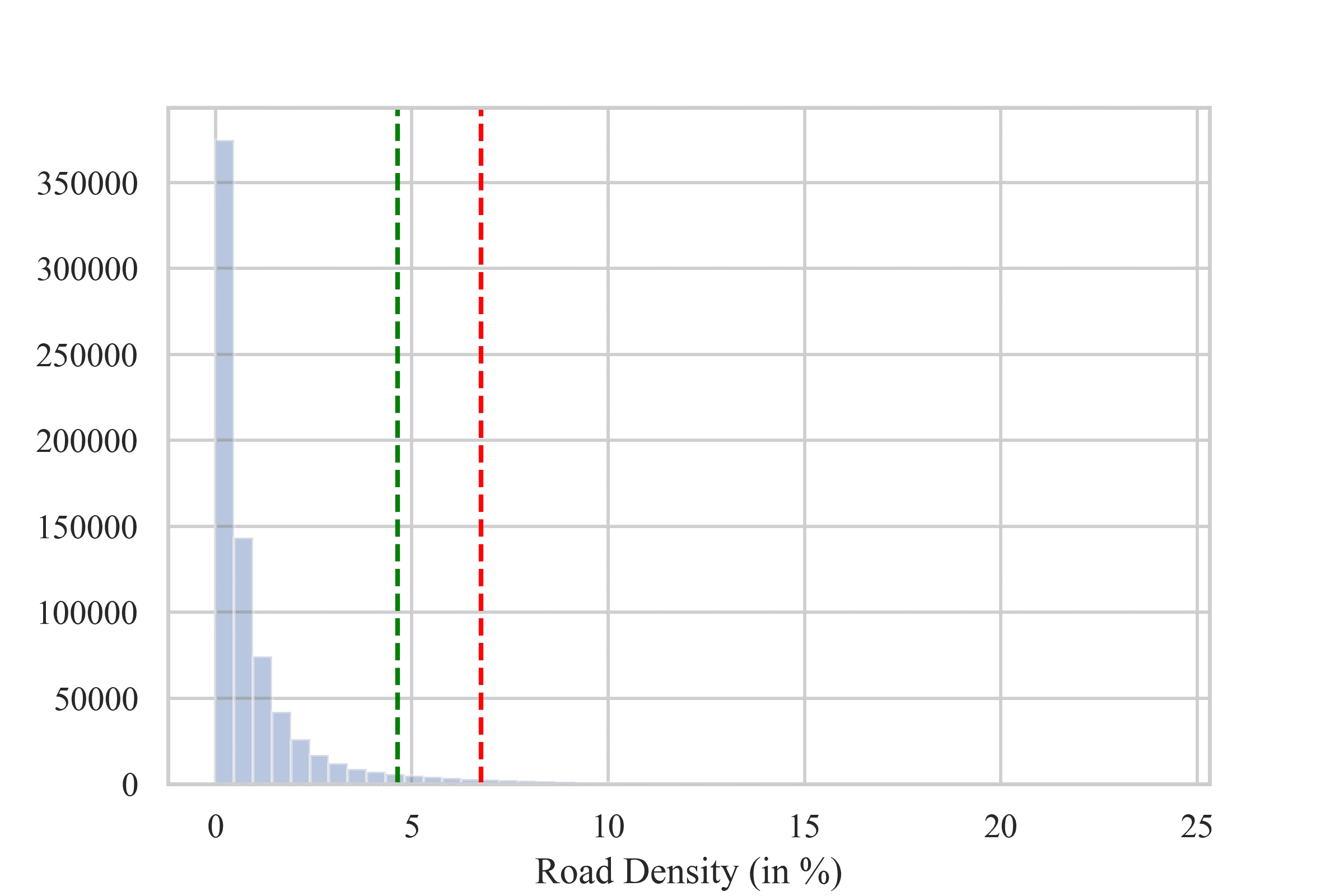}\hspace{5pt}
    \caption{The density of all the images we collect. The red dash line is the mean density of the DeepGlobe road extraction dataset while the green one is the Mnih Massachusetts roads dataset.}
    \label{fig:density_all}
\end{figure}

We define a term called \emph{road density} that is the ratio of the pixels of the road and the total pixels of a label.
The images and labels with the highest density are shown in Fig. \ref{fig:most}, and those with the lowest density are the area covered by forests, mountains, or the sea.
The density histogram of all the images we collect is shown in Fig. \ref{fig:density_all}.
We can find that most of the images are not covered by the roads or only contains very few roads, which are not suitable for pre-training.
The number of the images we use is based on the size of the DeepGlobe dataset as a baseline, which means that the number of pixels in $1\times$ dataset is the same as that in the DeepGlobe dataset.
We also make $2\times$, $5\times$, $10\times$, $20\times$, $50\times$ and $100\times$ in the similar way.


\subsection{Network Architecture}
\label{sec:network}
To test the versatility of the method, we tested a deeper U-Net (39.50M parameters, 8.76 GFLOPs), D-LinkNet with ResNet-18 (20.99M parameters, 23.88 GFLOPs), ResNet-34 (31.10M parameters, 33.56 GFLOPs), and ResNet-50 (217.65M parameters, 120.18 GFLOPs) (He et al. 2016; Zhou, Zhang and Wu 2018) as backbones.


\subsection{Why Are Images Stored in A Much Larger Size?}
\label{subsec:random_sample}
Since there are a massive number of images in our pre-training step, an effective caching method is critical. 
As mentioned in Section \ref{sec:aerial}, we stitch 256 tiles into a $4096\times4096$ image.
In the training step, we would load several large images into the memory, and there is no need to do the I/O very frequently because it would be more time-consuming to load the small but many files.
Moreover, we would randomly crop a patch from the large image as a data augmentation technique to improve the model's generalization. 
For example, the input size of the network is $512\times512$, and the size of the large image is $4096\times4096$. 
There are over 12 million \footnote {$(4096-512)^2 = 12,845,056$} possible input if the patch is cropped randomly, making the inputs from the large image different every iteration during training.





\begin{figure*}
    \centering
    \includegraphics[width=0.9\textwidth]{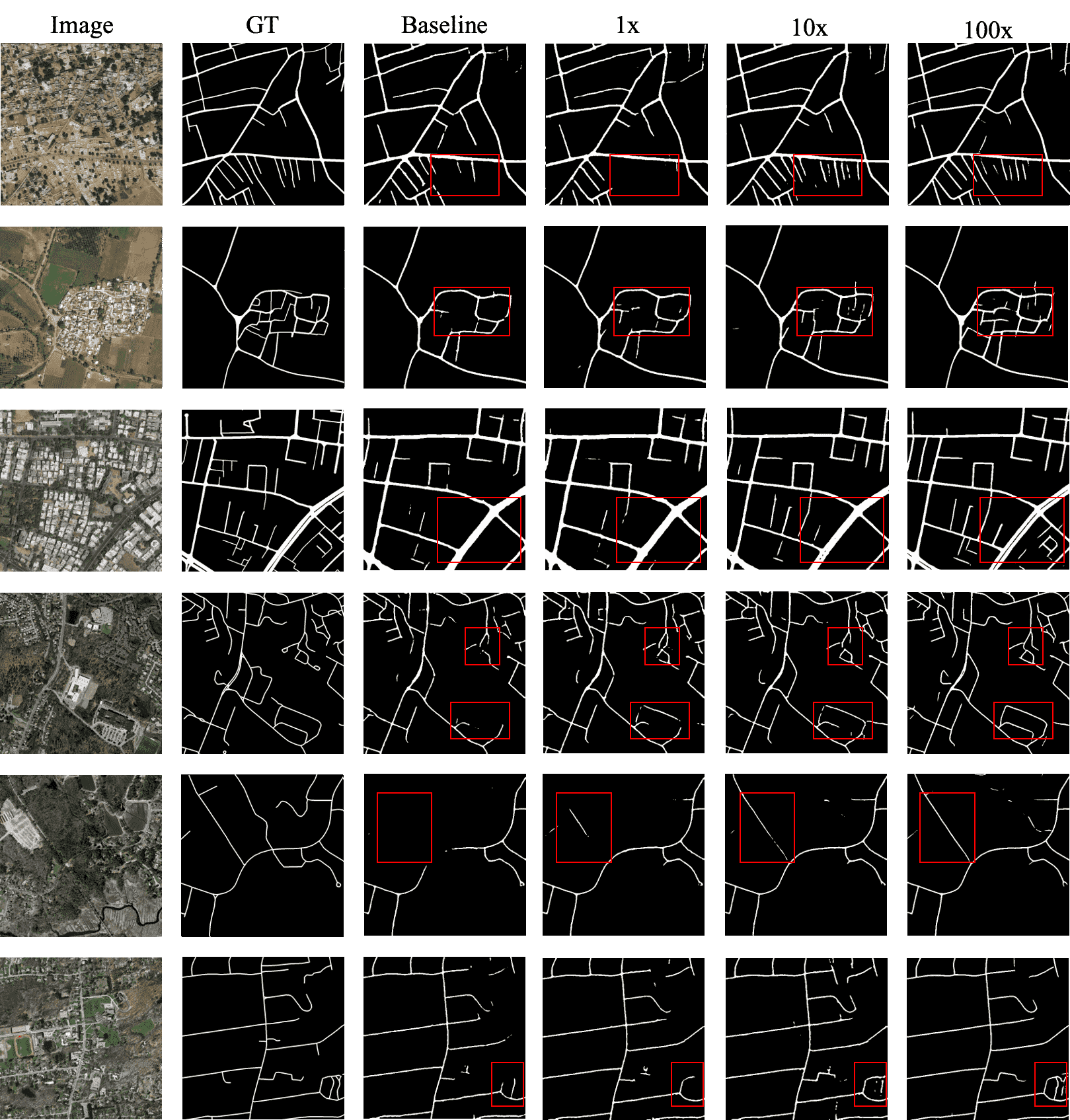}
    \caption{The visualization of the results using different size of pre-training dataset. The upper 3 rows show the example results of the DeepGlobe dataset, while the lower 3 are the results of the Mass dataset. From left to right is the image in the testing set, the ground truth of the corresponding image, the predicted mask of the baseline model, model pre-trained with $1\times$, $10\times$ and $100\times$ datasets. The results show that as the size of the pre-trained OSM dataset used for pre-training increases, the prediction results of the details in the results are better, as shown in the red boxes, which is consistent with the improvement of the mIoU indicator in the experiment.}
    \label{fig:results_size_vis}
\end{figure*}

\section{Experiments and Results}
\label{sec:exp}
\subsection{Datasets and evaluation}
We tested our results on the DeepGlobe road extraction (Demir et al. 2018) dataset (DeepGlobe dataset) and the Mnih Massachusetts roads (Mnih 2013) dataset (Mass dataset). 
The DeepGlobe dataset contains 5,603 images for training and 623 for validation, while the Mass dataset contains 1,107 images for training and 49 for validation. 
The \textit{density} of DeepGlobe dataset is 4.644\% while the Mass dataset is 6.764\%, which is shown in Fig. \ref{fig:density_all}.
The summary of the datasets we used is in Table \ref{tab:dataset_summary}.
The annotations in both datasets are finely labeled and only contain the background and road labels.
The evaluation metric we selected in experiments is mean intersection over union (mIoU) (Goodfellow et al. 2016).

\begin{table}[htbp]
\begin{center}
\begin{tabular}{lcc}
\hline
Dataset & Pixels No. & Mean Density (\%) \\
\hline
      Mass & $2.601\times10^{9}$  & $6.764$ \\
      DeepGlobe & $6.528\times10^{9}$  & $4.664$  \\
      Ours ($100\times$) & $6.529\times10^{11}$  & $6.566$ \\
      Ours (All) & $1.244\times10^{13}$  & $1.015$  \\
\hline
\end{tabular}
\end{center}
\caption{The size (number of pixels) and the mean density of the datasets we use}
\label{tab:dataset_summary} 
\end{table}

\subsection{Implementation Details}
Following Facebook's methods (He, Girshick and Doll{\'a}r 2019), we kept other settings the same except the pre-trained weight.
First, we trained our model on different scales of OSM datasets.
Then, we used the pre-trained weight to fine-tune the model on the target dataset to be tested.
The settings are the same both in the pre-training and fine-tuning phases. 
We trained our models on GTX 1080Ti GPUs, and the input image size of the model is $512\times512$. 
The batch size of every GPU is 6, and the learning rate is adjusted linearly based on the batch size according to (Mahajan et al. 2018).
The optimizer we use is Adam optimizer (Kingma and Ba 2014).
The loss function we use is dice soft loss, which is calculated as 
\begin{equation}
    DiceLoss = 1-\frac{2|Y_{pred} \bigcap Y_{gt}|+smooth}{|Y_{pred}|+|Y_{gt}|+smooth}
\end{equation}
where the smooth factor we use is 1.

During training, the largest learning rate is $1.6\times10^{-4}$ on 8 GPUs, while the least learning rate is $4\times10^{-5}$ on 2 GPUs.
We added an early-stop technique to prevent over-fitting.
If the mIoU did not increase over 6 epochs, we would decrease the current learning by 10 times.
If the learning rate is less than $10^{-7}$, the training process would be stopped. 
The settings of pre-training and fine-tuning are almost the same except for the number of GPUs, learning rate, and the training data.
For all the experiments, the $\emph{baseline}$ means that we train the network only use the target datasets which are the DeepGlobe dataset and the Mass dataset.
We show part of the visualization of the predicted results and the complete visualization can be found in the supplementary material.

\subsection{The Effect of Pre-training Size}
\label{sec:size}
\begin{figure}[htbp]
    \centering
    \includegraphics[width=0.8\columnwidth]{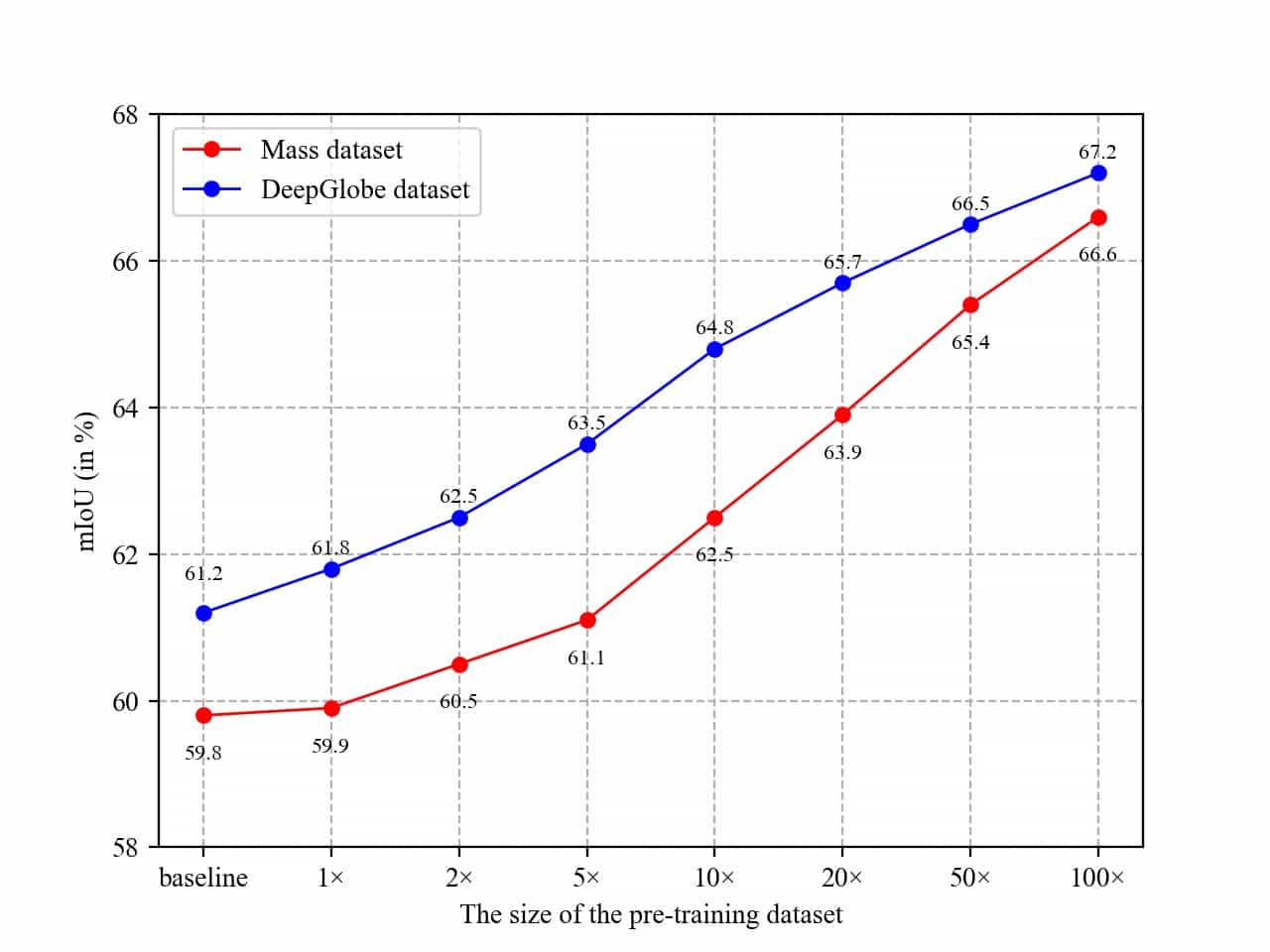}
    \caption{The mIoU of D-LinkNet with a ResNet-34 backbone, which is pre-trained using datasets with a mean density of 6.563\%, on the DeepGlobe dataset and the Mass dataset.}
    \label{fig:size}
\end{figure}

First, we investigate the impact of the pre-training size. 
Due to the limitation of our machines, we cannot use all the data we collect. 
We expand the size as large as possible and it is 100 times of the size of DeepGlobe and we compare the performance of 7 scales of the pre-training dataset from 1 time to 100 times.
The density of all 7 datasets is 6.563\%.
We choose D-LinkNet 34 because of its capacity and the target datasets and the results are in Fig. \ref{fig:size}.
As the size of the pre-training dataset increases, the mIoU becomes higher, which means that the larger the pre-training dataset, the better the model's performance after migration learning. 
The visualization of the results is shown in Fig. \ref{fig:results_size_vis}. 



\subsection{The Effect of Road Density}
\begin{figure}[htbp]
    \centering
    \subfigure[]{
        \includegraphics[width=0.45\columnwidth]{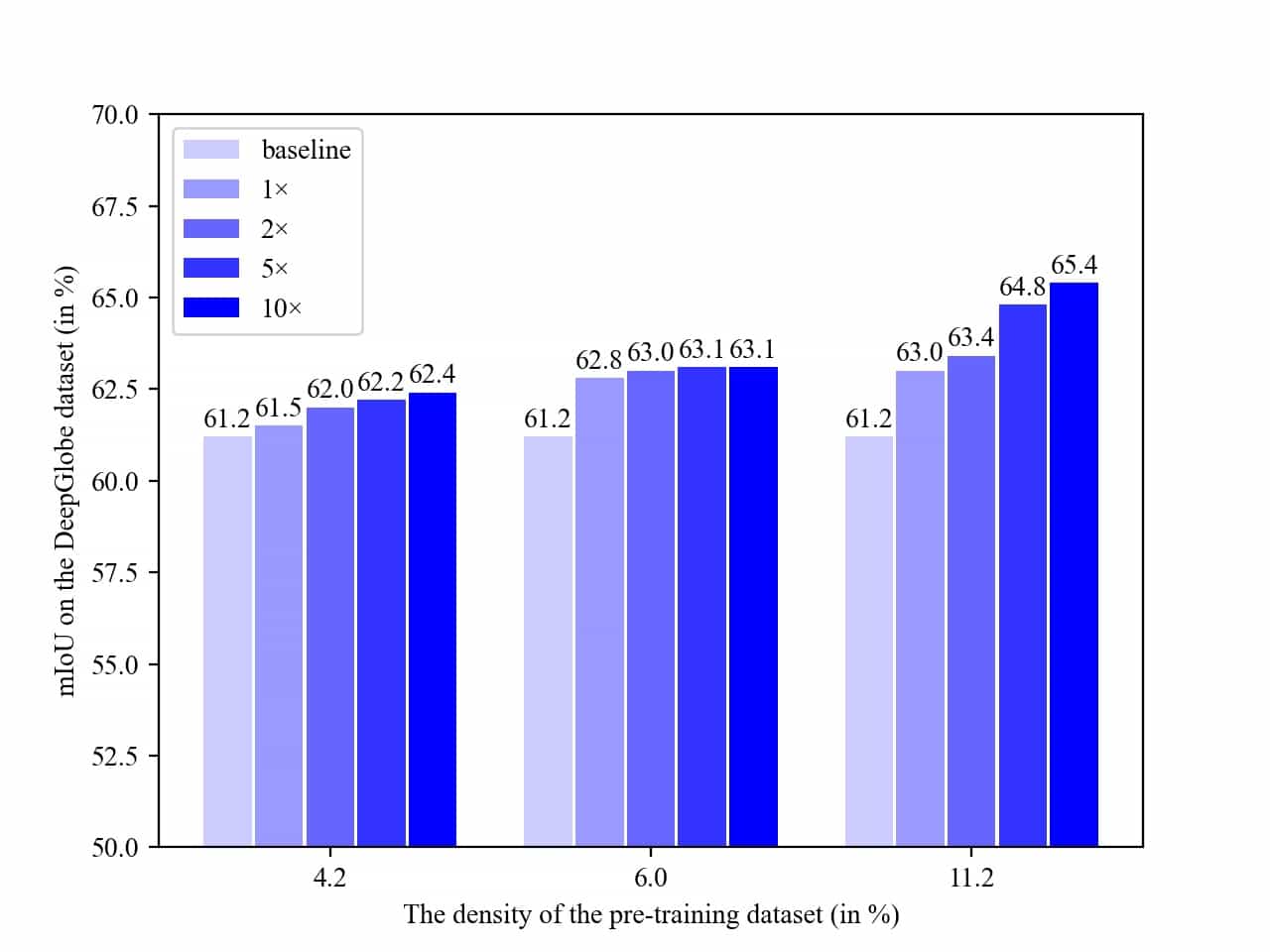}

    }
    \quad
    \subfigure[]{
	\includegraphics[width=0.45\columnwidth]{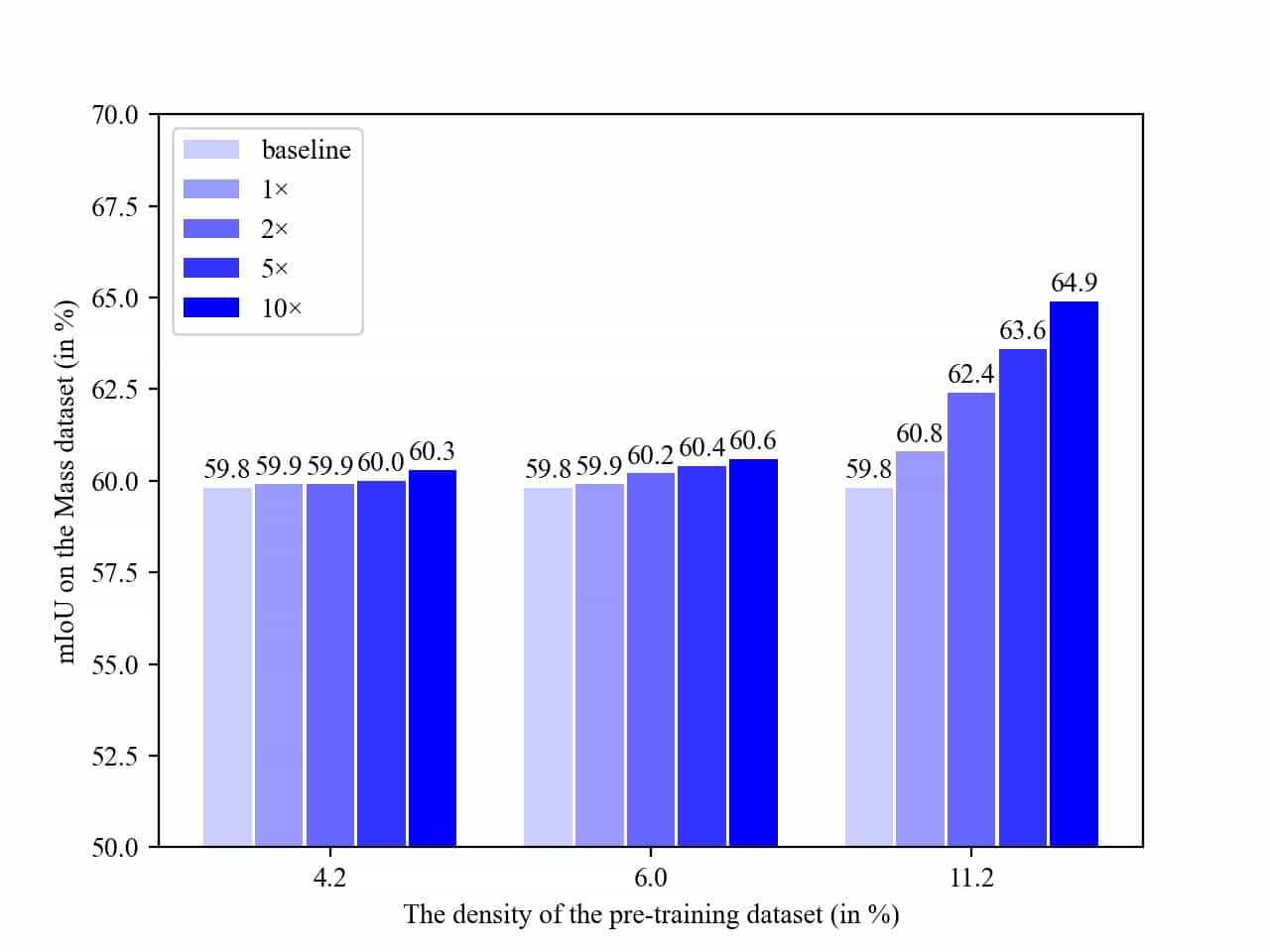}
    }

    \caption{The mIoU of D-LinkNet with a ResNet-34 backbone, which is pre-trained using datasets with a mean density of 6.563\%,on the DeepGlobe dataset in Fig. \ref{fig:density_exp}(a) and the Mass dataset in Fig. \ref{fig:density_exp}(b).}
\label{fig:density_exp}
\end{figure}

In addition to the size, road mean density is an important parameter in the dataset.
We use D-LinkNet 34 as the model and test the performance on the datasets with 3 different mean densities. 
The result is shown in Fig. \ref{fig:density_exp}.
We found that when the mean density of the pre-training dataset is relatively small, the mIoU only improves little with the size of the pre-training dataset increase.
No matter how high the mean density is, the mIoU of both datasets are the same when the pre-training datasets are as large as the DeepGlobe dataset. 
Meanwhile, it does not help a lot when the mean density of the pre-training dataset is the same or smaller than the target dataset.
It can be found that there is no obvious improvement on the Mass dataset, which mean density is 6.746\%, after pre-trained on datasets which mean densities are 4.2\% and 6.0\%.
The improvement is relatively small on the the DeepGlobe dataset, which mean density is 4.644\%, when the mean density of pre-training dataset is 4.2\%.
However, there is a significant improvement when the mean density of the pre-training dataset increased to 6.0\%
For the highest mean density, the improvement is obvious as the size of the pre-training dataset rises.

\subsection{The Effect of Fine-tuning Methods}
Our method is a kind of transfer learning, and we compare the performance using different feature transfer methods from the source dataset to the target one.
We design the $\emph{full network fine-tuning}$, $\emph{decoder fine-tuning}$ and $\emph{linear layer fine-tuning}$. All three methods would first load the pre-trained weight. For $\emph{full network fine-tuning}$, all the weight of the encoder and decoder would be trained.
For $\emph{decoder fine-tuning}$, the weight of the encoder would be frozen, and the weight of the decoder would be trained. 
For $\emph{linear layer fine-tuning}$, only the last 3 convolution layers of the decoder would be trained. The rest of the weight would be frozen during the training process.

\begin{figure}[htbp]
    \centering
    \subfigure[]{
        \includegraphics[width=0.45\columnwidth]{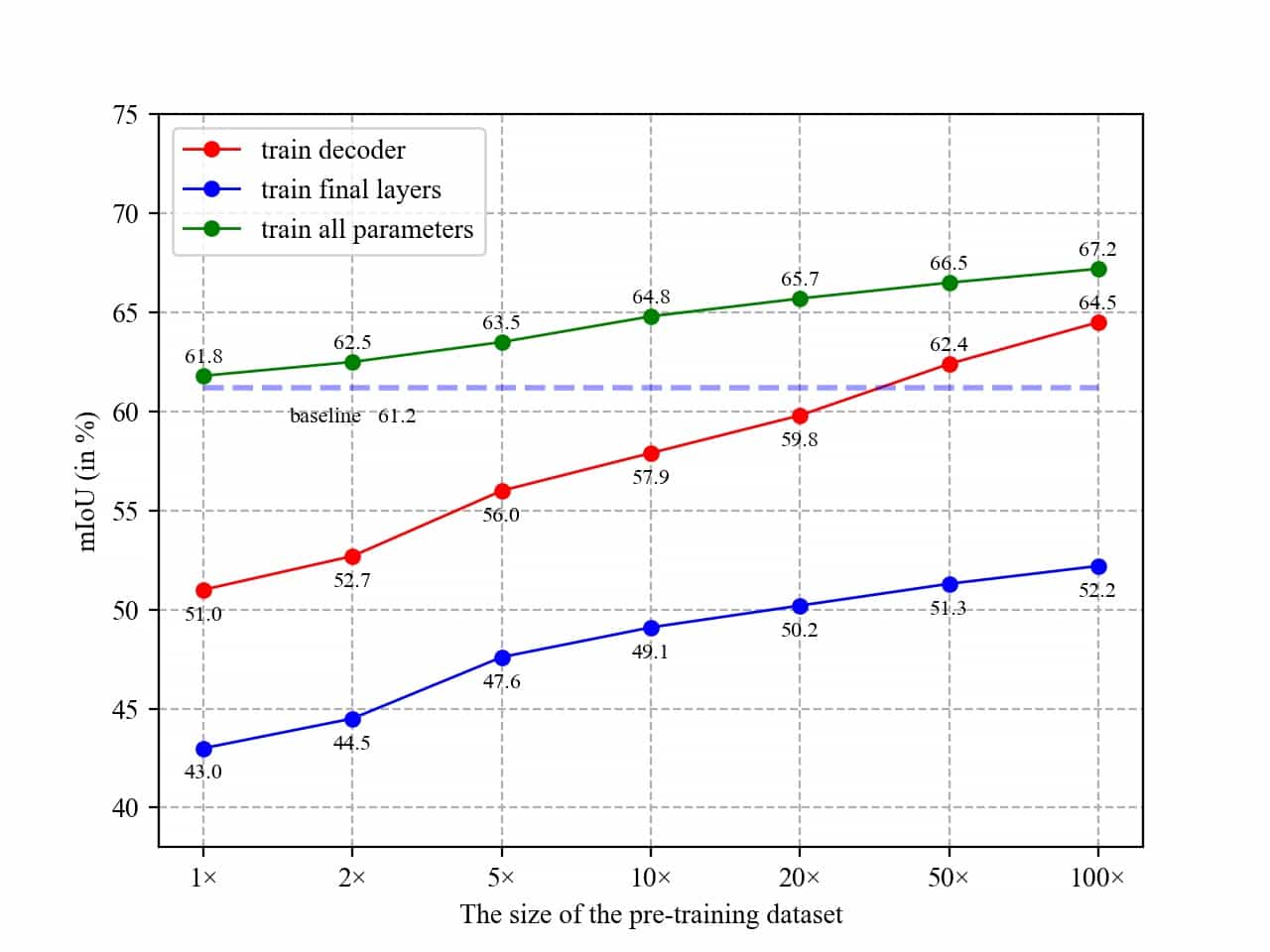}

    }
    \quad
    \subfigure[]{
	\includegraphics[width=0.45\columnwidth]{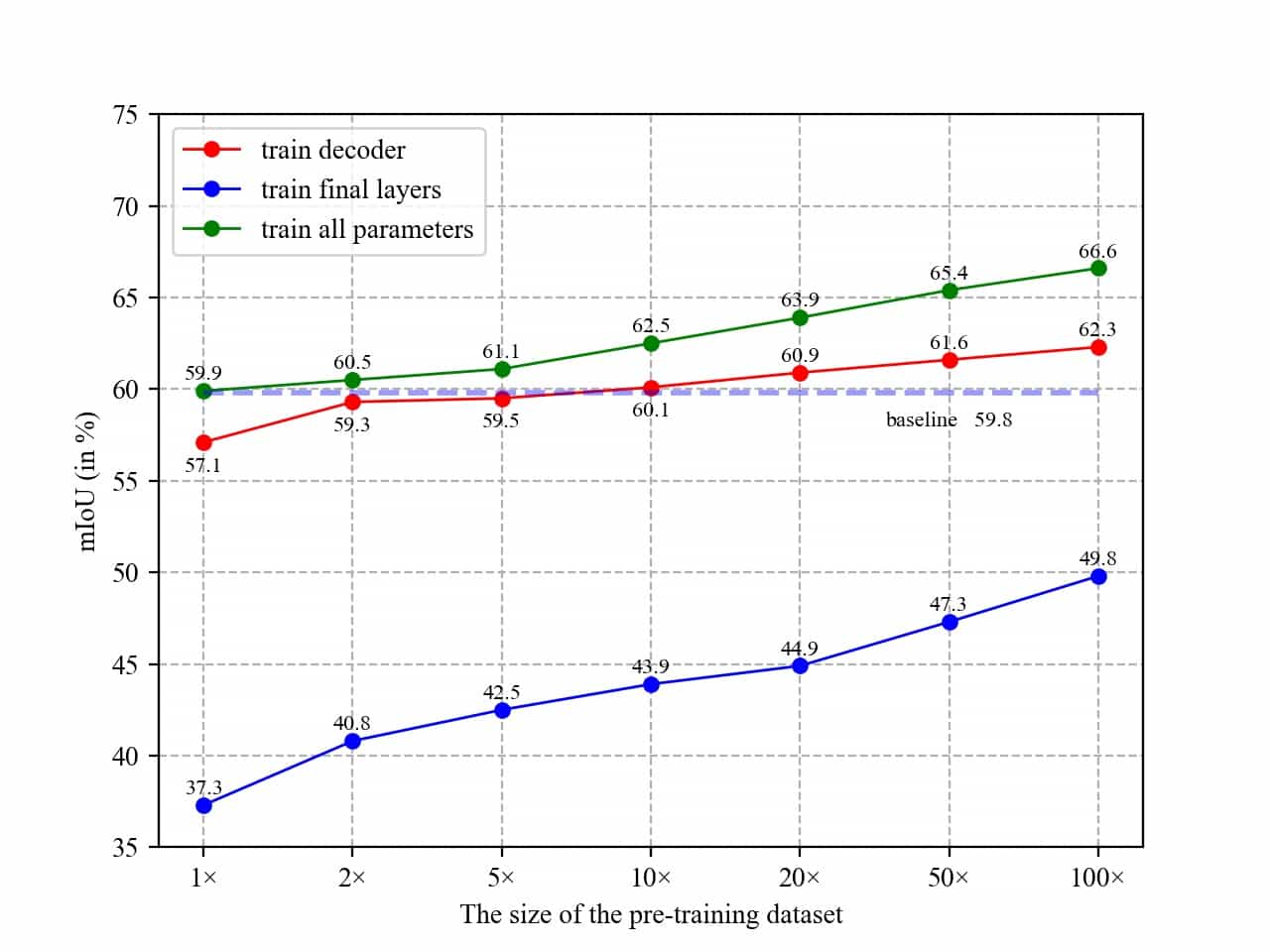}
    }

    \caption{The mIoU of D-LinkNet with a ResNet-34 backbone, which parameters are partly fixed during fine-tuning,on the DeepGlobe dataset in Fig. \ref{fig:freeze}(a) and the Mass dataset in Fig. \ref{fig:freeze}(b).}
\label{fig:freeze}
\end{figure}

We also compare the performance of the three fine-tuning methods on 7 datasets which is the same as in Section \ref{sec:size}, and the model is D-LinkNet with a ResNet-34 backbone.
The results are shown in Fig. \ref{fig:freeze}.
The model that only trains the final layers performs worst, even lower than the baseline result on every dataset.
The model which fixes the encoder can outperform the baseline when the scale of the pre-training dataset is big enough.
These two results demonstrate a large gap between the label generated by OSM and the fine-grained label, and it is necessary to train the whole network.
It also can be found that all the performance is positively correlated with the scale of the pre-training dataset, showing the effect of the size of the pre-training dataset although there is some noise.



\subsection{The Effect of Model Size}

To find the versatility of the method, we try 4 models as mentioned in Section \ref{sec:network}  using the ${100\times}$ dataset whose road density is 6.563\%.
We can find that the mIoU would be improved when a model with a higher capacity is used.
All the results are better than the baseline, showing the effectiveness of our method.
Because of the small size of the Mass dataset and the high capacity of ResNet-50, the D-LinkNet 50 is overfitting on the Mass dataset. 
Thus, the performance of the D-LinkNet 50 is not better than the D-LinkNet 34 on the Mass dataset.

\begin{figure}
    \centering
    \includegraphics[width=0.8\columnwidth]{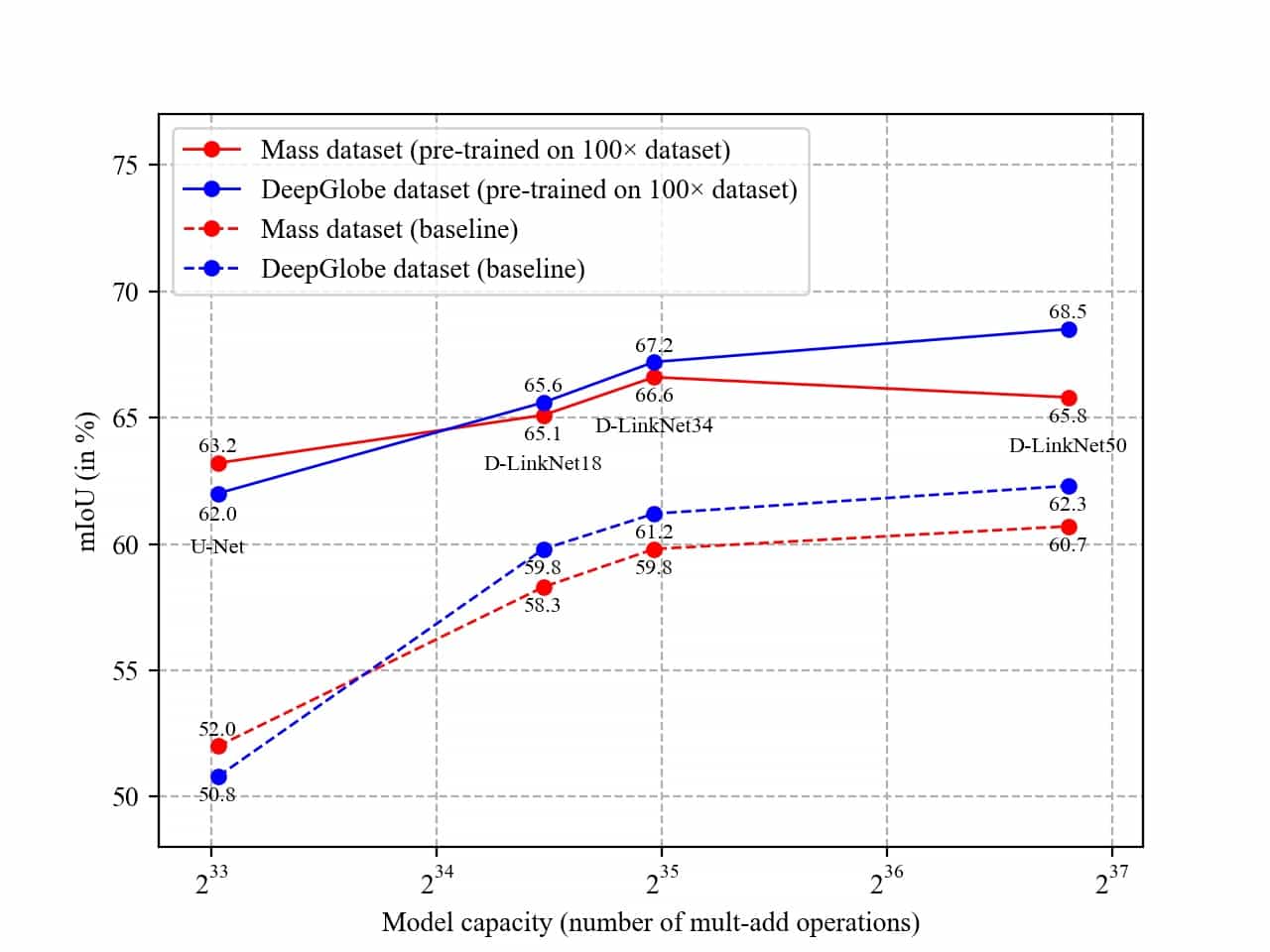}
    \caption{The mIoU of 4 models, which is pre-trained using the ${100\times}$ dataset with a mean density of 6.563\%, on the DeepGlobe dataset and the Mass dataset.}
    \label{fig:model_capacity}
\end{figure}

\subsection{Comparison With the Existing Method}
Since the release of the DeepGlobe dataset, many researchers use it as a baseline to evaluate their road extraction method. 
We also compare our methods with the existing road extraction methods on both datasets as shown in Table  \ref{tab:sota_deep} and our method outperforms the other methods.

\begin{table}[htbp]
\caption{The comparison with the existing methods}
\begin{center}
\begin{tabular}{lc}
\hline
Method & mIoU (\%) \\
\hline
      DeepRoadMapper (M{\'a}ttyus, Luo and Urtasun 2017) & 62.6  \\
      Topology Loss (Mosinska et al. 2018) & 64.9  \\
      D-LinkNet34 (Zhou, Zhang and Wu 2018) & 62.8 \\
      CADUNet (Li et al. 2021) & 62.1 \\
      Orientation (Batra et al. 2019) & 67.2  \\
      $\text { HG-Orientation }^{+} \text {-Attn }$ (Hamaguchi et al. 2021) & 68.3 \\
      \textbf{100$\times$ pre-trained D-LinkNet50 (Ours)} & \textbf{68.5} \\
\hline
\end{tabular}
\end{center}

\label{tab:sota_deep} 
\end{table}

\section{Conclusion and Future Work}
\label{sec:con}
In this paper, we propose a weakly supervised framework by utilizing the OSM data to make inaccurate labels, pre-train the model and transfer the pre-trained feature to the target dataset.
We find that the transfer learning performance would be improved when the size of the pre-training dataset is increased.
In addition to the size, the performance would also be largely influenced by the road density.
When the mean density of pre-training dataset is lower than the target dataset, the increment of size would help little.
However, if the mean density is relatively higher, the performance would increase on a large scale when the pre-training dataset becomes larger.
We find that the best method to perform transfer learning is to fine-tune the full network.
We also proved the versatility of our method regardless of the network framework and found that the network's performance is positively correlated with its capacity. 
Our work can provide a low-cost and fast method to improve performance consistently.

For future work, we plan to extend similar methods to building extraction because we have seen a considerable increase in accuracy in road extraction.
We also want to combine road extraction with contrastive learning, making the usage of the label more effective.


\begin{thebibliography}{}

\bibitem{bastani2018roadtracer}
Bastani, F., S. He, S. Abbar, M. Alizadeh, H. Balakrishnan, S. Chawla, S. Madden, and D. DeWitt. 2018. ``Roadtracer: Automatic extraction of road networks from aerial images.'' \emph{Paper presented at the Proceedings of the IEEE Conference on Computer Vision and Pattern Recognition.} Salt Lake City, UT, USA. June 18-21.

\bibitem{batra2019improved}
Batra, A., S. Singh, G. Pang, S. Basu, C. Jawahar, and M. Paluri. 2019. ``Improved road connectivity by joint learning of orientation and segmentation.'' \emph{Paper presented at the Proceedings of the IEEE/CVF Conference on Computer Vision and Pattern Recognition.} Long Beach, CA, USA. June 15-21.

\bibitem{bonafilia2019building}
Bonafilia, D., J. Gill, S. Basu, and D. Yang. 2019. ``Building high resolution maps for humanitarian aid and development with weakly-and semi-supervised learning,'' \emph{Paper presented at the Proceedings of the IEEE/CVF Conference on Computer Vision and Pattern Recognition Workshops.} Long Beach, CA, USA. June 15-21.

\bibitem{buslaev2018fully}
Buslaev, A., S. Seferbekov, V. Iglovikov, and A. Shvets. 2018. ``Fully convolutional network for automatic road extraction from satellite imagery,'' \emph{Paper presented at the Proceedings of the IEEE Conference on Computer Vision and Pattern Recognition Workshops.} Salt Lake City, UT, USA. June 18-21.

\bibitem{DeepGlobe18}
Demir, I., K. Koperski, D. Lindenbaum, G. Pang, J. Huang, S. Basu, F. Hughes, D. Tuia, and R. Raskar. 2018. ``Deepglobe 2018: A challenge to parse the earth through satellite images.'' \emph{Paper presented at the Proceedings of the IEEE Conference on Computer Vision and Pattern Recognition Workshops.} Salt Lake City, UT, USA. June 18-21.

\bibitem{goodfellow2016deep}
Goodfellow, I., Y. Bengio, and A. Courville. 2016. \emph{Deep learning}: MIT press.

\bibitem{osm}
Haklay, M., and P. Weber. 2008. ``Openstreetmap: User-generated street maps.'' \emph{IEEE Pervasive computing} 7 (4):12-8.

\bibitem{hamaguchi2021heterogeneous}
Hamaguchi, R., Y. Furukawa, M. Onishi, and K. Sakurada. 2021. ``Heterogeneous Grid Convolution for Adaptive, Efficient, and Controllable Computation.'' \emph{Paper presented at the Proceedings of the IEEE/CVF Conference on Computer Vision and Pattern Recognition.} June 19-25.

\bibitem{he2016deep}
He, K., X. Zhang, S. Ren, and J. Sun. 2016. ``Deep residual learning for image
  recognition,'' \emph{Paper presented at the Proceedings of the IEEE conference on computer vision and pattern recognition.} Las Vegas, NV, USA. June 26-30.

\bibitem{he2019rethinking}
He, K., R. Girshick, and P. Dollár. 2019. ``Rethinking imagenet pre-training,'' \emph{Paper presented at the Proceedings of the IEEE Conference on Computer Vision and Pattern Recognition Workshops.} Long Beach, CA, USA. June 15-21.

\bibitem{kingma2014adam}
Kingma, D. P., and J. Ba. 2014. ``Adam: A method for stochastic optimization.'' \emph{arXiv preprint arXiv:1412.6980.}

\bibitem{krizhevsky2012imagenet}
Krizhevsky, A., I. Sutskever, and G. E. Hinton. 2017. ``ImageNet classification with deep convolutional neural networks.'' \emph{Commun. ACM} 60 (6):84–90. doi: 10.1145/3065386.

\bibitem{li2021cascaded}
Li, J., Y. Liu, Y. Zhang, and Y. Zhang. 2021. ``Cascaded Attention DenseUNet (CADUNet) for Road Extraction from Very-High-Resolution Images.'' \emph{ISPRS International Journal of Geo-Information} 10 (5):329.

\bibitem{mattyus2017deeproadmapper}
Máttyus, G., W. Luo, and R. Urtasun. 2017. ``Deeproadmapper: Extracting road
  topology from aerial images,'' \emph{Paper presented at the Proceedings of the IEEE International Conference on Computer Vision.} Venice, Italy. October 22-29.
  
\bibitem{mahajan2018exploring}
Mahajan, D., R. Girshick, V. Ramanathan, K. He, M. Paluri, Y. Li, A. Bharambe, and L. V. D. Maaten. 2018. ``Exploring the limits of weakly supervised
  pretraining,'' \emph{Paper presented at the Proceedings of the European conference on computer vision (ECCV).} Munich, Germany. September 8–14.

\bibitem{mnih2012learning}
Mnih, V., and G. E. Hinton. 2012. ``Learning to label aerial images from noisy data.'' \emph{Paper presented at the Proceedings of the 29th International conference on machine learning (ICML-12).} Edinburgh, Scotland. June 26 – July 1.

\bibitem{mass}
Mnih V. 2013. ``Mnih massachusetts roads dataset.'' http://www.cs.toronto.edu/~vmnih/data/.

\bibitem{mosinska2018beyond}
Mosinska, A., P. Marquez-Neila, M. Koziński, and P. Fua. 2018. ``Beyond the pixel-wise loss for topology-aware delineation.'' \emph{Paper presented at the Proceedings of the IEEE conference on computer vision and pattern recognition.}

\bibitem{pan2021generic}
Pan, D., M. Zhang, and B. Zhang. 2021. ``A Generic FCN-Based Approach for the Road-Network Extraction From VHR Remote Sensing Images–Using OpenStreetMap as Benchmarks.'' \emph{IEEE Journal of Selected Topics in Applied Earth Observations and Remote Sensing} 14:2662-73.

\bibitem{ronneberger2015u}
Ronneberger, O., P. Fischer, and T. Brox. 2015. ``U-net: Convolutional networks for
  biomedical image segmentation,'' \emph{Paper presented at the International Conference on Medical image computing and computer-assisted intervention.} Munich, Germany. October 5-9.

\bibitem{shan2020cross}
Shan, B., and Y. Fang. 2020. ``A Cross Entropy Based Deep Neural Network Model for Road Extraction from Satellite Images.'' \emph{Entropy} 22 (5):535.

\bibitem{sun2018combining}
Sun, T., Z. Di, and Y. Wang. 2018. ``Combining Satellite Imagery and GPS Data for Road Extraction.'' \emph{In Proceedings of the 2nd ACM SIGSPATIAL International Workshop on AI for Geographic Knowledge Discovery}, 29–32. Seattle, WA, USA: Association for Computing Machinery.

\bibitem{sun2019leveraging}
Sun, T., Z. Di, P. Che, C. Liu, and Y. Wang. 2019. ``Leveraging crowdsourced gps data for road extraction from aerial imagery.'' \emph{Paper presented at the Proceedings of the IEEE/CVF Conference on Computer Vision and Pattern Recognition.} Long Beach, CA, USA. June 15-21.

\bibitem{wan2021roadnet}
Wan, J., Z. Xie, Y. Xu, S. Chen, and Q. Qiu. 2021. ``DA-RoadNet: A Dual-Attention Network for Road Extraction from High Resolution Satellite Imagery.'' \emph{IEEE Journal of Selected Topics in Applied Earth Observations and Remote Sensing} 14:6302-6315.

\bibitem{wang2021nl}
Wang, Y., J. Seo, and T. Jeon. 2021. ``NL-LinkNet: Toward lighter but more accurate road extraction with nonlocal operations.'' \emph{IEEE Geoscience and Remote Sensing Letters.} 19 (1):1-5.

\bibitem{wu2019road}
Wu, S., C. Du, H. Chen, Y. Xu, N. Guo, and N, Jing. 2021. ``Road extraction from very high resolution images using weakly labeled OpenStreetMap centerline.'' \emph{ISPRS International Journal of Geo-Information.} 8 (11):478.

\bibitem{zhang2018road}
Zhang, Z., Q. Liu, and Y. Wang. 2018. ``Road extraction by deep residual u-net.'' \emph{IEEE Geoscience and Remote Sensing Letters} 15 (5):749-53.

\bibitem{zhang2019aerial}
Zhang, X., X. Han, C. Li, X. Tang, H. Zhou, and L. Jiao. 2019. ``Aerial Image Road Extraction Based on an Improved Generative Adversarial Network.'' \emph{Remote Sensing} 11 (8):930.

\bibitem{zhou2018d}
Zhou, L., C. Zhang, and M. Wu. 2018. ``D-linknet: Linknet with pretrained encoder and
  dilated convolution for high resolution satellite imagery road extraction,'' \emph{Paper presented at the Proceedings of the IEEE Conference on Computer Vision and Pattern Recognition Workshops.} Salt Lake City, UT, USA. June 18-21.

\bibitem{zoph2020rethinking}
Zoph, B., G. Ghiasi, T. Lin, Y. Cui, H. Liu, E. D. Cubuk, and Q. V. Le. 2020. ``Rethinking pre-training and self-training,'' \emph{arXiv preprint
  arXiv:2006.06882}.
  

  
  
  
  
 








  
  










\end{thebibliography}
\end{document}